\documentclass[10pt,twocolumn,letterpaper]{article}
\usepackage{authblk}
\usepackage[pagenumbers]{cvpr}
\newcommand{\papertitle}{DRIC-Agent: Drift-Controlled Visual Benchmark Construction from Sparse Specifications}

\usepackage[T1]{fontenc}
\usepackage{algorithm}
\usepackage{algpseudocode}
\usepackage{array}
\definecolor{cvprblue}{rgb}{0.21,0.49,0.74}
\usepackage[breaklinks,colorlinks,allcolors=cvprblue]{hyperref}
\usepackage[capitalize]{cleveref}
\title{\papertitle}

\author[1]{Langxi Huang}
\author[1]{Pingping Zhang}
\author[2]{Lanyun Zhu}
\author[3]{Chunyang Jiang}
\author[1]{Jiawei Shao}
\author[1]{Haocheng Yuan}
\author[1]{Peilin Chen}
\author[1]{Shiqi Wang}
\affil[1]{City University of Hong Kong}
\affil[2]{Tongji University}
\affil[3]{Hong Kong University of Science and Technology}
\date{}
\hypersetup{
  pdftitle={DRIC-Agent: Drift-Controlled Visual Benchmark Construction from Sparse Specifications},
  pdfauthor={Langxi Huang, Pingping Zhang, Lanyun Zhu, Chunyang Jiang, Jiawei Shao, Haocheng Yuan, Peilin Chen, Shiqi Wang}
}
\begin{document}
\maketitle

\begin{abstract}
As multimodal models evolve, existing visual benchmarks may not adequately assess newly identified capability gaps, creating a need for new, targeted benchmarks. When these benchmarks are constructed automatically, however, the generated images and questions may fail to test the intended capability. To address this diagnostic drift, we introduce \textbf{DRIC-Agent}, a multi-agent framework for \textbf{DRI}ft-\textbf{C}ontrolled visual benchmark construction from sparse specifications. Its core is an adaptive controller that diagnoses failures against the original specification to select the level and scope of correction across generated content, evaluators, and derived requirements. Requirement revisions are reviewed against the originally specified capability and difficulty. Under unchanged requirements and evaluators, data repairs are accepted only when they resolve failed checks without introducing new failures. Following a requirement or evaluator revision, the controller re-evaluates the current data to establish a new validation baseline, distinguishing content improvement from changes in assessment. The 15 evaluated MLLMs achieve accuracies ranging from 55.5\% to 91.0\%. Complete target fidelity is 73.7\% versus 47.4\% for direct construction. Human evaluation achieves 86.0\% answer validity, 83.5\% diagnostic target preservation, and 99.6\% visual judgment necessity, supporting the effectiveness of DRIC-Agent in constructing diagnostically faithful visual benchmarks.
\end{abstract}

\section{Introduction}

\label{sec:intro}

As multimodal large language models (MLLMs) evolve, existing visual benchmarks may not adequately assess newly identified capability gaps. Examining these gaps calls for targeted benchmarks designed around specific perceptual or reasoning abilities. Constructing such benchmarks requires translating evaluation goals into suitable visual content, questions, reference answers, and validation procedures.
Automated benchmark construction seeks to coordinate these activities, making it possible to generate targeted tests while systematically checking their quality.

Recent benchmark-construction frameworks improve reliability by organizing generation and verification around explicit construction requirements. Structured plans, verification feedback, and shared specifications provide mechanisms for checking generated artifacts and correcting defects in content, evaluation code, or the specifications themselves~\cite{jiang2026embodiedbenchclaw, ivanov2026anchor}. These mechanisms help establish consistency between generated artifacts and the requirements used to assess them. However, consistency with the current requirements does not by itself establish fidelity to the original evaluation goal: the requirements may incompletely or incorrectly capture the capability to be assessed.

This distinction matters when the input specifies the capability to assess but leaves the system to derive the requirements for individual test samples. The system must translate that goal into concrete decisions about visual evidence, question formulation, and validation. During this translation, inferred requirements may omit or alter an aspect of the intended capability. If these requirements guide both generation and verification, a sample may satisfy its checks while failing to assess the original target. We call this deviation from the intended capability during test construction diagnostic drift. Correcting this drift requires reviewing both the generated content and the requirements used to generate and validate it. Evidence must then guide whether to revise the content, the checking procedures, or the requirements themselves, with proposed requirement changes assessed against the original diagnostic goal.

To address this diagnostic drift, we introduce \textbf{DRIC-Agent}, a multi-agent framework for \textbf{DRI}ft-\textbf{C}ontrolled visual benchmark construction from sparse specifications. Its core is an adaptive controller that determines the scope of correction and the conditions for accepting each revision. For data validation, evaluators supply computational or semantic observations, while fixed runtime rules compare these observations with the targets specified by the reviewed requirements. When validation reveals a mismatch, LLM-based critics jointly examine the candidate data, evaluator implementations, and the source clauses underlying the requirements to propose a diagnosis. The controller uses this diagnosis to direct correction to the relevant agents, subject to explicit acceptance rules. With requirements and evaluators unchanged, a data repair is accepted only when it resolves failed checks without introducing new failures. Revising a requirement instead requires a source-based review that identifies a defect and assesses the proposed change against the originally specified capability and difficulty; repeated failures may trigger this review but do not themselves authorize a revision. Changes to requirements or evaluators trigger renewed validation of the affected data, with evaluator bindings rebuilt where necessary, before construction proceeds.

Our contributions are threefold:
\begin{itemize}
    \item We formulate visual benchmark construction from sparse specifications as preserving a diagnostic target while deriving and revising test content and evaluation requirements.

    \item We introduce DRIC-Agent, whose adaptive controller uses source reviews and execution evidence to determine whether to correct generated content, evaluators, or inferred requirements.

    \item We define acceptance rules that require data repairs to reduce failures without introducing new ones, and re-establish the validation baseline after requirements or evaluators change.
\end{itemize}

We evaluate 290 samples across 15 MLLMs, examine natural construction deviations, compare 19 paired constructions, and assess human judgments and targeted follow-up samples.

\section{Related work}
\label{sec:related}

\subsection{Benchmark validity and item-level evidence}
Benchmarks should measure the capabilities they are intended to assess, yet construct misalignment remains common~\cite{bean2025constructvalidity}. ECBD connects evaluation claims to task evidence~\cite{liu2024ecbd}; Datasheets and BenchmarkCards document assumptions underlying datasets and benchmarks~\cite{gebru2021datasheets,sokol2025benchmarkcards}. BenchBrowser and item-level releases enable inspection beyond aggregate scores~\cite{diddee2026benchbrowser,jiang2026itemlevel}; Targeted Tests records construction and audit decisions~\cite{li2026targetedtests}. MLLM-as-a-Judge identifies biases and inconsistent judgments, motivating human validation of automated assessments~\cite{chen2024mllmjudge}. DRIC-Agent applies these concerns to inferred requirements and generated artifacts, retaining evidence of their review and revision.

\begin{table}[!t]
\caption{Comparison of benchmark construction methods. SI: synthesis without sample-level source assets; Tgt.: explicit capability/error target; Req.: target-linked validation of derived requirements; Sem.: semantic artifact checks; Tr.: decision-linked construction trace. $\checkmark$, $\triangle$, and -- denote reported, partial, and unreported support. $^\dagger$Non-VQA.}
\label{tab:construction-authority}
\centering
\footnotesize
\setlength{\tabcolsep}{2.3pt}
\begin{tabular*}{\columnwidth}{@{\extracolsep{\fill}}lccccc@{}}
\toprule
Method & SI & Tgt. & Req. & Sem. & Tr. \\
\midrule
AutoBencher$^\dagger$~\citep{li2025autobencher} & $\triangle$ & $\triangle$ & -- & $\triangle$ & $\triangle$ \\
AutoBench-V~\citep{bao2024autobenchv} & $\checkmark$ & $\checkmark$ & -- & $\checkmark$ & $\triangle$ \\
ArenaBencher$^\dagger$~\citep{chen2025arenabencher} & -- & $\checkmark$ & -- & $\checkmark$ & -- \\
ReachQA~\citep{he2025reachqa} & -- & $\triangle$ & -- & $\checkmark$ & -- \\
ChartGen~\citep{kondic2025chartgen} & -- & -- & -- & -- & -- \\
ChartM$^3$~\citep{wang2025chartm3} & -- & $\triangle$ & -- & $\checkmark$ & -- \\
BloomQA$^\dagger$~\citep{chen2026bloomqa} & -- & $\checkmark$ & $\checkmark$ & $\triangle$ & -- \\
ChartVerse~\citep{liu2026chartverse} & $\checkmark$ & -- & -- & $\checkmark$ & -- \\
ChartNet~\citep{liu2026chartnet} & -- & $\triangle$ & -- & $\triangle$ & -- \\
CoSyn~\citep{yang2025cosyn} & $\checkmark$ & $\triangle$ & -- & -- & -- \\
Benchmark Agent~\citep{li2026benchmarkagent} & -- & $\checkmark$ & $\checkmark$ & $\checkmark$ & -- \\
RankJudge$^\dagger$~\citep{tang2026rankjudge} & -- & $\checkmark$ & $\checkmark$ & $\checkmark$ & $\triangle$ \\
ProbeLLM$^\dagger$~\citep{huang2026probellm} & -- & $\triangle$ & -- & $\checkmark$ & $\triangle$ \\
Embodied-BenchClaw~\citep{jiang2026embodiedbenchclaw} & -- & $\checkmark$ & $\triangle$ & $\checkmark$ & $\checkmark$ \\
Anchor$^\dagger$~\citep{ivanov2026anchor} & -- & $\triangle$ & $\triangle$ & $\checkmark$ & $\triangle$ \\
PIPE-Cypher$^\dagger$~\citep{ranganath2026pipecypher} & -- & $\triangle$ & $\triangle$ & $\checkmark$ & $\checkmark$ \\
Targeted Tests$^\dagger$~\citep{li2026targetedtests} & -- & $\checkmark$ & $\triangle$ & $\checkmark$ & $\checkmark$ \\
\midrule
\textbf{DRIC-Agent} & $\checkmark$ & $\checkmark$ & $\checkmark$ & $\checkmark$ & $\checkmark$ \\
\bottomrule
\end{tabular*}

\end{table}

\begin{figure*}[t]
\centering
\includegraphics[width=\textwidth]{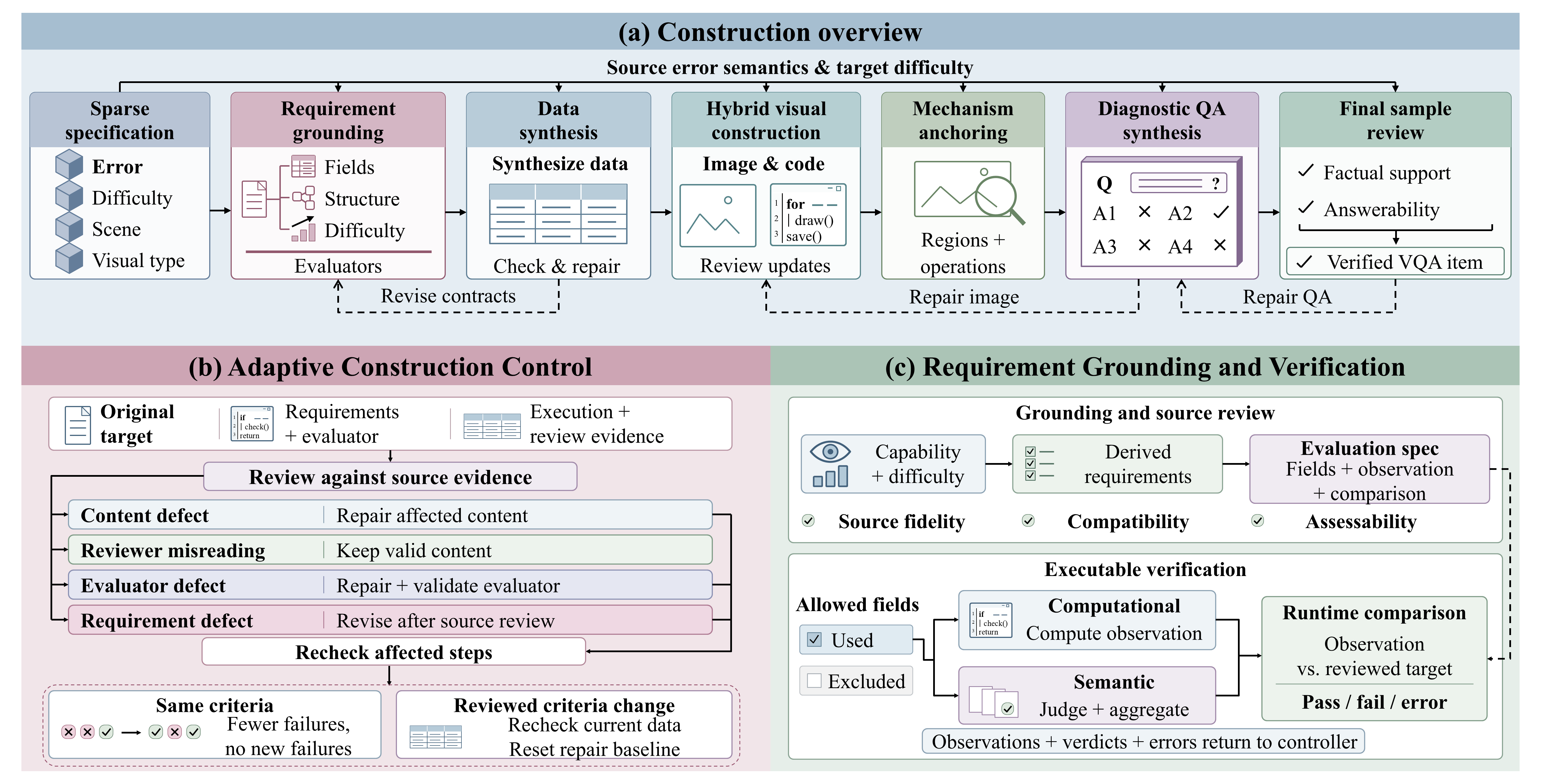}
\caption{\textbf{Drift-controlled visual benchmark construction.} \textbf{(a)} The error-taxonomy target and difficulty guide construction from requirements and data to visual evidence and diagnostic QA. Dashed arrows indicate repair or backtracking. \textbf{(b)} Source-based diagnosis distinguishes content defects, reviewer misreadings, evaluator defects, and requirement defects, determining what may change. Under fixed requirements and evaluators, accepted data repairs must reduce recorded failures without introducing new ones; authorized assessment changes require re-evaluation and a new baseline. This rule governs check-level progress, not guaranteed semantic improvement. \textbf{(c)} Reviewed requirements define permitted fields, observations, and comparisons. Computational or semantic evaluators supply observations; runtime rules compare them with reviewed targets to produce verdicts. Observations, verdicts, and errors return to the controller; errors do not count as passes. Requirement revisions must preserve the original capability and difficulty, and affected downstream checks must be rerun. The construction trace retains versions, evidence, and decisions for inspection.}
\label{fig:overview}
\end{figure*}

\begin{figure*}[t]
\centering
\begin{minipage}[t]{0.54\textwidth}
\vspace{0pt}\centering
{\small\bfseries (a) Heatmap}\par\vspace{2pt}
\includegraphics[width=\linewidth]{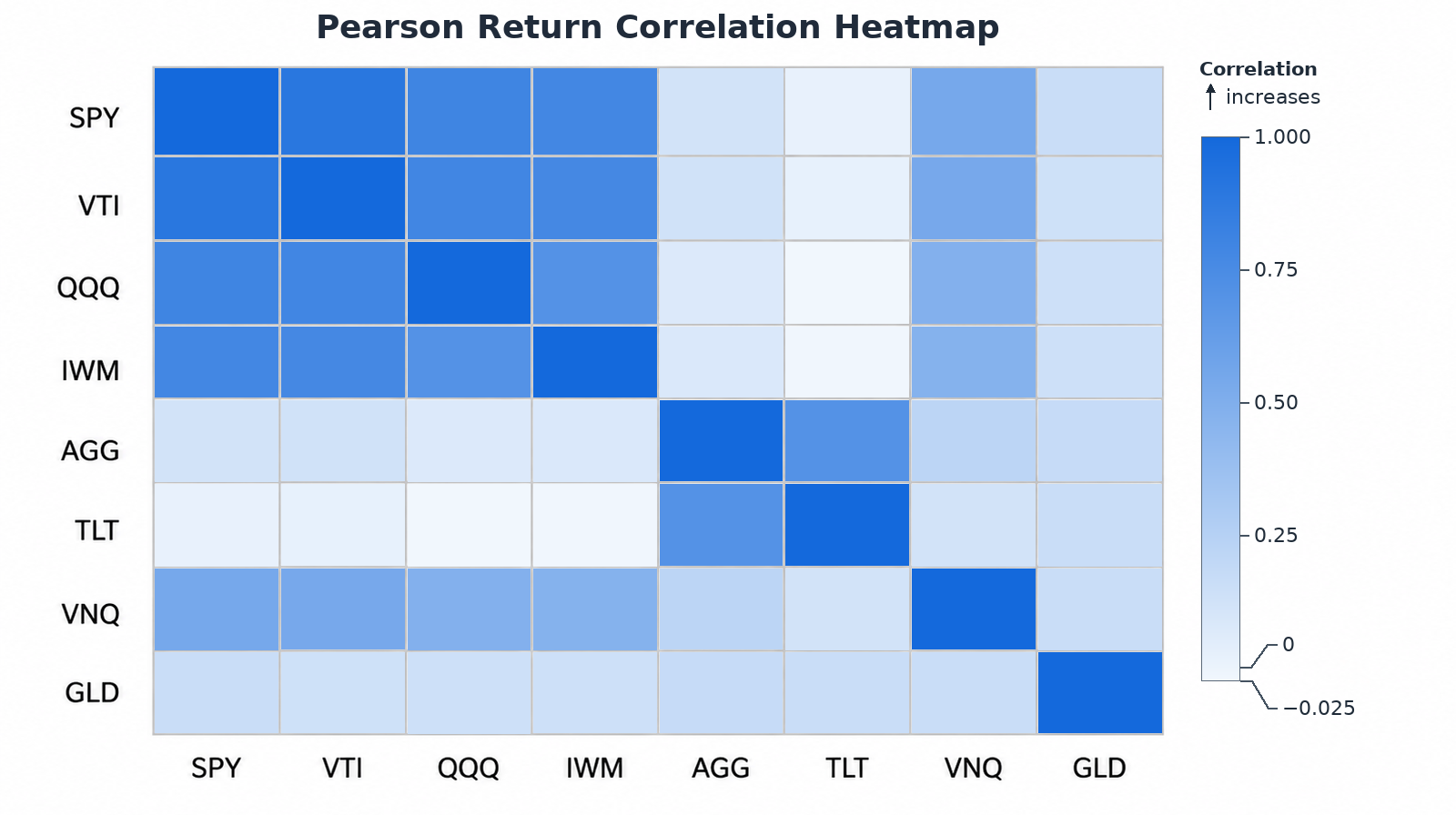}
\end{minipage}\hfill
\begin{minipage}[t]{0.44\textwidth}
\vspace{0pt}\normalsize\raggedright
\textbf{Question.} According to the heatmap's color scale, which of the listed assets has a correlation with IWM strictly between 0.50 and 0.75? Explain your choice by comparing the saturation of each listed asset's IWM cell with the 0.50 and 0.75 scale levels, including why each alternative falls outside the interval.
\par\vspace{6pt}
\renewcommand{\arraystretch}{1.25}
\begin{tabular}{@{}lc@{}}
A. VNQ & \textcolor{red!80!black}{$\times$} \\
B. QQQ & \textcolor{green!50!black}{$\checkmark$} \\
C. VTI & \textcolor{red!80!black}{$\times$} \\
D. SPY & \textcolor{red!80!black}{$\times$} \\
\end{tabular}
\end{minipage}
\par\vspace{7pt}
\begin{minipage}[t]{0.24\textwidth}\vspace{0pt}\centering
{\footnotesize\bfseries (b) Map}\par\vspace{2pt}
\includegraphics[width=\linewidth]{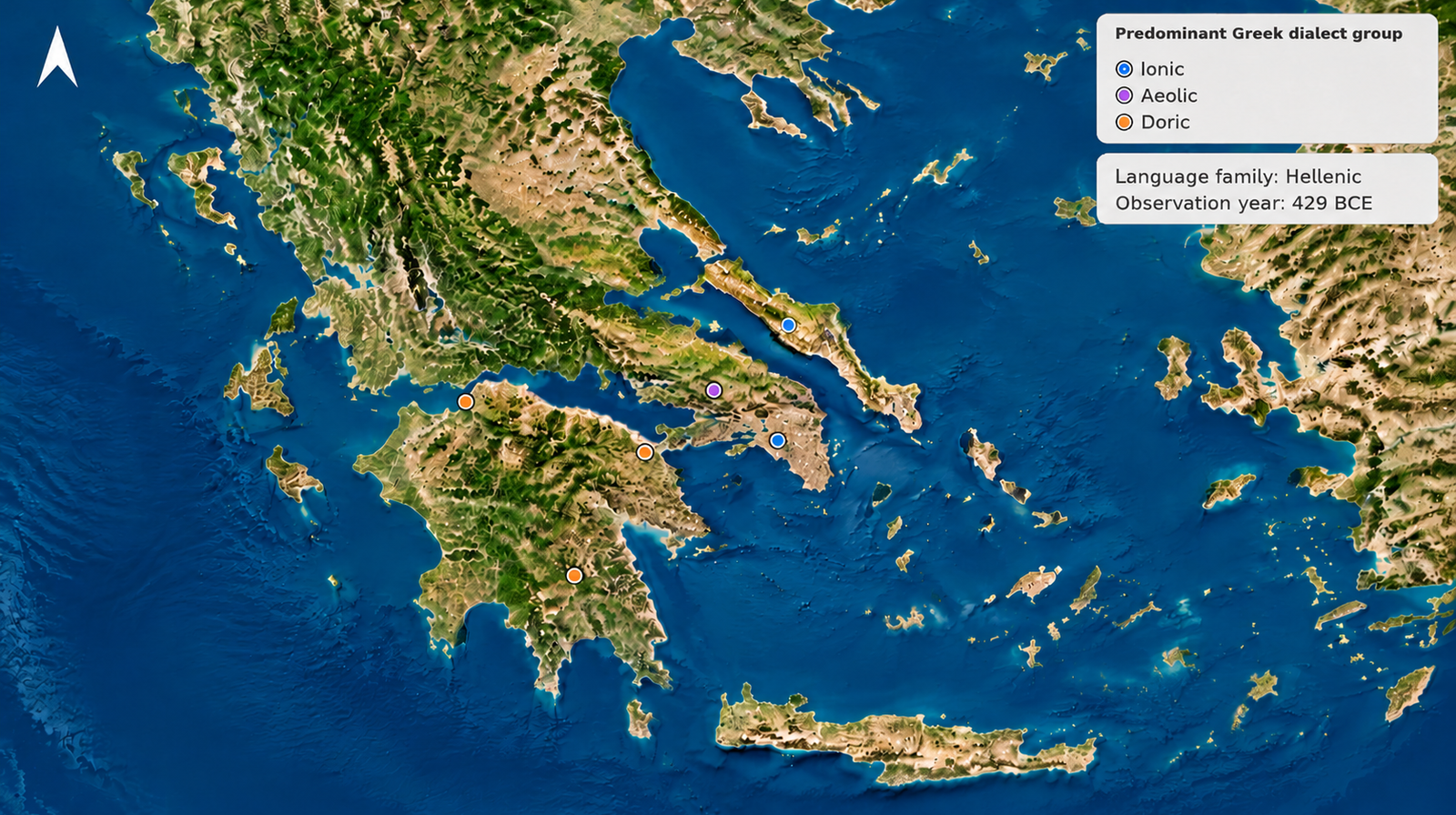}
\end{minipage}\hfill
\begin{minipage}[t]{0.24\textwidth}\vspace{0pt}\centering
{\footnotesize\bfseries (c) Comic page}\par\vspace{2pt}
\includegraphics[width=\linewidth]{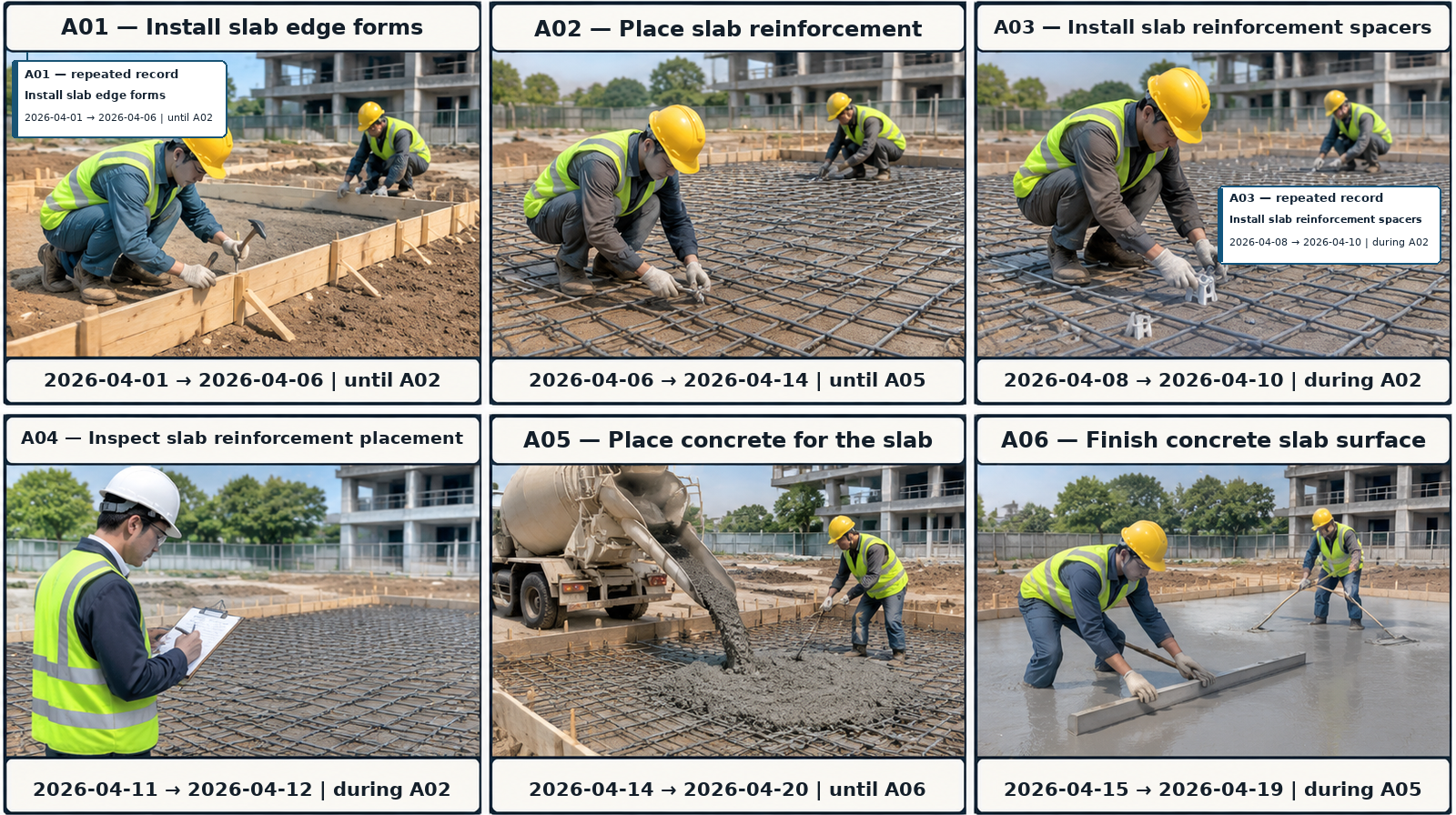}
\end{minipage}\hfill
\begin{minipage}[t]{0.24\textwidth}\vspace{0pt}\centering
{\footnotesize\bfseries (d) Pictorial instruction}\par\vspace{2pt}
\includegraphics[width=\linewidth]{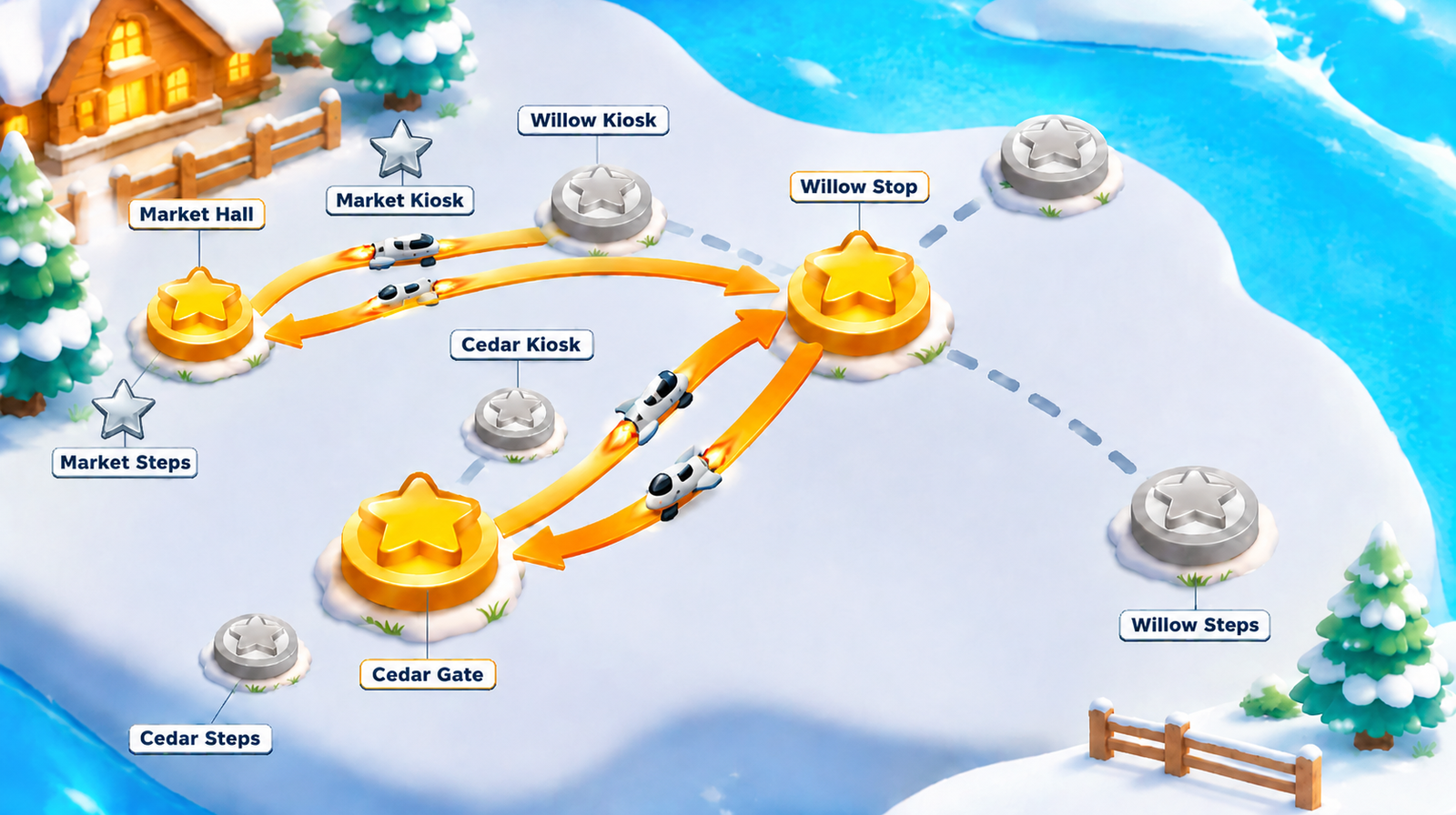}
\end{minipage}\hfill
\begin{minipage}[t]{0.24\textwidth}\vspace{0pt}\centering
{\footnotesize\bfseries (e) Box plot}\par\vspace{2pt}
\includegraphics[width=\linewidth]{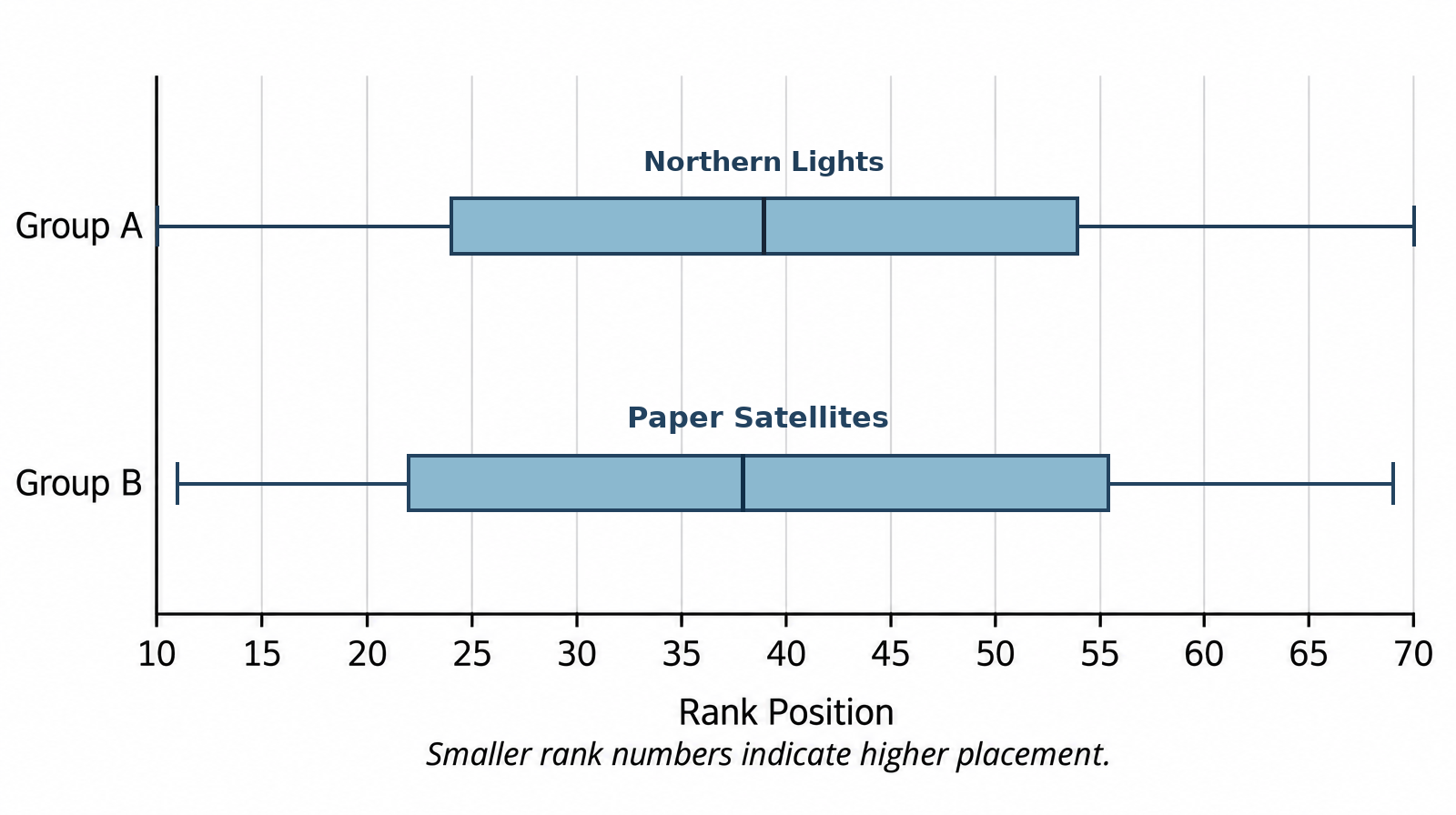}
\end{minipage}
\caption{\textbf{Diagnostic samples across diverse visual types.} \textbf{(a)} The question requires locating asset correlations on the heatmap's color scale. B is the reference answer; 9 of 15 evaluated models answered incorrectly, while all four human reviewers endorsed answer validity, target conditions, and judgment necessity. \textbf{(b--e)} Geographic maps, construction sequences, illustrated instructions, and box plots show the range of visual evidence supported by the workflow.}
\label{fig:visual-examples}
\end{figure*}

\subsection{Automated benchmark construction}
Task Me Anything generates tailored tests from a taxonomy of visual assets~\cite{zhang2024taskmeanything}. BenchAgents plans generation and verification~\cite{butt2024automatedbenchmark}; Benchmark Agent grounds planned subtasks in retrieved data~\cite{li2026benchmarkagent}. AutoBencher selects topics using model feedback~\cite{li2025autobencher}; ArenaBencher evolves items around evaluation targets~\cite{chen2025arenabencher}. BloomQA derives verified tasks from expert guidelines~\cite{chen2026bloomqa}. AutoHallusion manipulates images to probe hallucination~\cite{sun2024autohallusion}. RankJudge verifies dialogue flaws against documents and plans~\cite{tang2026rankjudge}; ProbeLLM generates verifiable tests to identify recurring errors~\cite{huang2026probellm}.

Feedback refinement~\cite{dhuliawala2024chainverification,madaan2023selfrefine,shinn2023reflexion}, structured decoding~\cite{geng2023grammar,geng2025jsonschema}, and policy checks~\cite{simbola2026compliance} support validation. PIPE-Cypher combines schema grounding, execution, and semantic review~\cite{ranganath2026pipecypher}. Embodied-BenchClaw controls drift through expert-backed contracts and local repair~\cite{jiang2026embodiedbenchclaw}; Anchor revisits workflow specifications when verification reveals defects~\cite{ivanov2026anchor}. DRIC-Agent reviews requirements inferred during construction and authorizes corrections against the original diagnostic target.

\subsection{Visual data and VQA synthesis}
CLEVR annotates required reasoning while reducing answer biases~\cite{johnson2017clevr}; GQA controls question semantics and answer distributions through scene graphs and functional programs~\cite{hudson2019gqa}. ChartGen reconstructs charts into code for augmentation~\cite{kondic2025chartgen}; ChartNet scales the generation of paired images, data, and QA~\cite{liu2026chartnet}. ReachQA generates questions from plotting code~\cite{he2025reachqa}; ChartM$^3$ combines templates, code generation, and multimodal verification~\cite{wang2025chartm3}. CoSyn produces text-rich visuals and QA from code~\cite{yang2025cosyn}; ChartVerse checks generated questions against source-code answers~\cite{liu2026chartverse}. AutoBench-V expands capability specifications, checks image consistency, and reduces text-only answerability~\cite{bao2024autobenchv}. DRIC-Agent checks whether visual evidence and required judgments preserve the error-taxonomy target. Table~\ref{tab:construction-authority} compares construction methods.

\section{Method}
\label{sec:method}

DRIC-Agent constructs visual tests from sparse diagnostic specifications. Its adaptive controller uses execution evidence and reviews against the original specification to guide revisions to generated content, evaluators, and inferred requirements. Explicit acceptance rules determine which proposed revisions are adopted. We first define the task, then present the controller, requirement grounding and verification, and diagnostic sample synthesis.

\subsection{Problem formulation}
\label{sec:problem_formulation}

Given a sparse specification $s=\langle c,v,\tau,d\rangle$, our task is to construct a visual test of the capability implicated by the specified error-taxonomy entry $\tau$ at the requested difficulty $d$. The specification also provides an initial scene $c$ and visual type $v$, but no sample-level materials or acceptance criteria. The system must therefore derive both the test content and its assessment requirements, while retaining $\tau$ and $d$ as the reference for subsequent revisions.

Successful construction produces
\begin{equation}
    y=\langle D,I,M,q,O,a,T\rangle.
    \label{eq:construction_output}
\end{equation}
The test comprises an image $I$, a question $q$, answer options $O$, and a reference answer $a$. Its supporting records include structured data $D$, mechanism anchors $M$ linking the target to visual evidence and intended operations, and a construction trace $T$. The evaluated model receives only $(I,q,O)$, which must provide sufficient information to answer the question.

For each specification, construction proceeds iteratively, starting from the specification and stage-specific budgets. At decision step $t$, the state $X_t$ records the current stage, adopted artifacts and tentative candidates, requirements and evaluators, accumulated evidence, and remaining budgets. The controller selects an action and updates the state as
\begin{equation}
\begin{aligned}
    u_t &= \pi(X_t;s),\\
    X_{t+1} &= \mathcal{U}\bigl(X_t,\operatorname{Execute}(X_t,u_t);s\bigr).
\end{aligned}
\label{eq:construction_iteration}
\end{equation}
Here $\pi$ selects a permitted action, $\operatorname{Execute}$ returns its outputs and evidence, and $\mathcal{U}$ applies the revision and acceptance rules. Each updated state informs the next decision for the same sample, enabling repeated generation, verification, and correction, including permitted backtracking. The process returns $y$ when the required final checks pass, or stops without an accepted sample when no permitted continuation remains within the budgets. Section~\ref{sec:adaptive_control} defines the action-selection and state-update rules.

\subsection{Adaptive construction control}
\label{sec:adaptive_control}

The controller implements $\pi$ and $\mathcal{U}$ in Equation~\ref{eq:construction_iteration} through evidence-based action selection and explicit acceptance rules. Content repair and changes to assessment conditions require different rules for evaluating progress.

\paragraph{Evidence and correction scope.}
Local failures prompt bounded correction under current requirements. Persistent data-check failures trigger joint review of source clauses, requirements, evaluator code, execution results, and data before and after repair. A data violation supports data repair; an implementation mismatch supports evaluator repair. A critic's misreading instead calls for reconsidering the finding while retaining valid content. Figure~\ref{fig:overview}(b) summarizes these diagnoses and their acceptance rules.

Requirement review does not itself authorize revision. Repeated failures can trigger review when the diagnoses permit escalation; an unresolved evaluator mismatch or unproven diagnosis does not authorize requirement changes. Review examines source fidelity, available evidence, and compatibility with other requirements, including those whose checks passed. Only identified defects justify revision, and proposed changes must pass review against the original capability and difficulty.

Section~\ref{sec:semantic-control} analyzes naturally occurring semantic deviations in retained construction records.

\paragraph{Content repair under fixed assessment conditions.}
Let $\nu$ denote the current assessment version, comprising the requirements, evaluators, and runtime evaluation rules. For data $D$, the recorded failed-check set is
\begin{equation}
    \mathcal{F}_{\nu}(D)=\{r\in R_D^{\nu}:V_r^{\nu}(D)=0\},
    \label{eq:failed_checks}
\end{equation}
where $R_D^{\nu}$ is the set of data requirements for version $\nu$ and $V_r^{\nu}$ is the verdict defined in Section~\ref{sec:requirements_verification}. At a controller step $t$ devoted to data repair, let $D_t$ be the adopted baseline and $D'_t$ the candidate. With $\nu$ unchanged, the data component of $\mathcal{U}$ follows
\begin{equation}
    D_{t+1}=
    \begin{cases}
        D'_t, & \mathcal{F}_{\nu}(D'_t)\subsetneq \mathcal{F}_{\nu}(D_t),\\
        D_t, & \text{otherwise},
    \end{cases}
    \label{eq:repair_progress}
\end{equation}
provided the candidate is structurally valid, evaluation completes, and previously failed requirements are re-evaluated; otherwise, $D_t$ is retained. Adoption therefore resolves at least one recorded failure without introducing another. After complete evaluation, a nonempty candidate that reduces the failure count relative to its working predecessor may be retained separately for further repair. It replaces the adopted baseline only under Equation~\ref{eq:repair_progress}. This progress rule applies to data repair; visual and question revisions use their stage-specific reviews. This is check-level monotonicity under fixed assessment conditions, not a guarantee of monotonic diagnostic fidelity; semantic fidelity still requires review against $s$.

\paragraph{Revision of assessment conditions.}
Evaluator corrections must pass implementation review; computational repairs are executed on failure data and reviewed with their results. Requirement corrections must pass source review. An authorized revision creates a new assessment version $\nu'$. Affected evaluator bindings are rebuilt, outdated verdicts are invalidated, and affected artifacts are re-evaluated. If data checks change, the current data are reassessed under $\nu'$ to establish a new repair baseline. Cross-version verdict changes are not counted as data-repair progress.

\paragraph{Continuation and backtracking.}
Candidates enter the adopted state only after satisfying the applicable acceptance rule; eligible partial results remain separate. Construction advances only when all required current-stage checks pass for the current artifacts and assessment version. Artifact revisions invalidate affected downstream verdicts; these checks must be rerun before their outputs are reused. Image defects found during anchor or question review can return construction to visual synthesis, while final question defects can return it to question construction. Backtracking restores required inputs, with data fixed throughout these stages. Every attempt updates evidence, trace $T$, and budget usage for the next decision, even without adoption. Semantic diagnoses remain fallible.

Algorithm~\ref{alg:construction} summarizes the control loop. Here $e_t$ contains the outputs and evidence returned by $\operatorname{Execute}$; $\nu$ and $D_t$ come from $X_t$, and $D'_t$ is a returned data candidate. The admitted data $D^{\star}$ are defined in Section~\ref{sec:sample_synthesis}. The branches specify $\mathcal U$ and do not assume infallible diagnoses.

\begin{algorithm}[t]
\caption{Drift-controlled construction}\label{alg:construction}
\small
\begin{algorithmic}[1]
\Require $s=\langle c,v,\tau,d\rangle$ and stage budgets
\Ensure Accepted $y$, or termination with trace $T$
\State Initialize $X_0$ from $s$; $T\gets()$; $t\gets0$
\While{a permitted action remains within budget}
    \State $u_t\gets\pi(X_t;s)$ under the control rules
    \State $e_t\gets\operatorname{Execute}(X_t,u_t)$
    \If{an assessment revision passes required reviews}
        \State Rebuild affected bindings; invalidate old verdicts
        \State Re-evaluate affected artifacts under $\nu'$
        \State Reset data baseline if data checks changed
    \ElsIf{data repair keeps assessment version $\nu$ fixed}
        \State Adopt $D'_t$ only under Eq.~\ref{eq:repair_progress}
    \Else
        \State Apply stage reviews; retain valid content if misread
    \EndIf
    \State Recheck affected outputs; backtrack as permitted
    \State Keep $D^{\star}$ fixed in visual and QA stages
    \State $X_{t+1}\gets\mathcal U(X_t,e_t;s)$; update $T$, budgets
    \If{all required final checks pass in $X_{t+1}$}
        \State \Return $y=\langle D,I,M,q,O,a,T\rangle$ from $X_{t+1}$
    \EndIf
    \State $t\gets t+1$
\EndWhile
\State \Return termination with $T$
\end{algorithmic}
\end{algorithm}

\begin{table*}[!t]
\caption{Model accuracy (\%) on 290 diagnostic samples. Top-level group sizes appear in the headers; Overall pools all samples. $\Delta_{\mathrm{L1}}$: highest minus lowest top-level group accuracy (percentage points). Bold: best within each model group. Group means weight models equally.}
\label{tab:model-accuracy}
\centering
\small
\setlength{\tabcolsep}{3pt}
\renewcommand{\arraystretch}{1.08}
\begin{tabular*}{\textwidth}{@{\extracolsep{\fill}}lcccccc@{}}
\toprule
Model & \shortstack{Overall\\($n=290$)} &
\shortstack{Visual Perception\\\& Encoding\\($n=80$)} &
\shortstack{Structure\\\& Layout\\($n=60$)} &
\shortstack{Data Logic\\\& Reasoning\\($n=82$)} &
\shortstack{Semantic\\Understanding\\($n=68$)} & $\Delta_{\mathrm{L1}}$ \\
\midrule
\multicolumn{7}{@{}l}{\textbf{Closed-source models}} \\
GPT-6.1-Sol & \textbf{91.0} & \textbf{93.8} & 88.3 & \textbf{95.1} & 85.3 & 9.8 \\
GPT-5.6-Sol & 83.8 & 82.5 & 85.0 & 87.8 & 79.4 & 8.4 \\
Gemini 4 Hydrogen & 89.7 & 87.5 & 88.3 & 92.7 & 89.7 & 5.2 \\
Gemini 4 Carbon & 90.3 & 90.0 & 86.7 & 92.7 & 91.2 & 6.0 \\
Gemini 3.8 Flash & 87.9 & 86.2 & \textbf{91.7} & 86.6 & 88.2 & 5.4 \\
Gemini 3.6 Flash & 83.8 & 77.5 & 86.7 & 85.4 & 86.8 & 9.3 \\
Claude Haiku 5.5 & 89.0 & 81.2 & \textbf{91.7} & 93.9 & 89.7 & 12.7 \\
Claude Sonnet 5.5 & 89.7 & 85.0 & 90.0 & 90.2 & \textbf{94.1} & 9.1 \\
Claude Opus 5.5 & 90.0 & 91.2 & \textbf{91.7} & 91.5 & 85.3 & 6.4 \\
Claude Opus 5 & 90.0 & 85.0 & 85.0 & \textbf{95.1} & \textbf{94.1} & 10.1 \\
\addlinespace[2pt]Group mean & 88.5 & 86.0 & 88.5 & 91.1 & 88.4 & -- \\
\midrule
\multicolumn{7}{@{}l}{\textbf{Open-source models}} \\
DeepSeek V4.1 Flash & 55.5 & 53.8 & 55.0 & 54.9 & 58.8 & 5.1 \\
GLM-5.3 & 83.8 & 76.2 & 85.0 & 91.5 & 82.4 & 15.2 \\
GLM-5.2 & \textbf{88.6} & \textbf{88.8} & \textbf{86.7} & \textbf{92.7} & \textbf{85.3} & 7.4 \\
Qwen3.8 Flash & 59.7 & 58.8 & 71.7 & 53.7 & 57.4 & 18.0 \\
Kimi K3 & 68.6 & 62.5 & 73.3 & 67.1 & 73.5 & 11.0 \\
\addlinespace[2pt]Group mean & 71.2 & 68.0 & 74.3 & 72.0 & 71.5 & -- \\
\bottomrule
\end{tabular*}

\end{table*}

\subsection{Requirement grounding and verification}
\label{sec:requirements_verification}

\paragraph{Grounding requirements in the diagnostic target.}
Compatibility screening first checks the initial scene and visual type, allowing adjustments while preserving the diagnostic target and requested difficulty. Specification agents then derive component requirements from the seed and accepted preceding outputs, expressing difficulty through component-specific conditions. Critics review source fidelity, compatibility, and assessability. Requirements that fail review are revised and reviewed again before governing synthesis. Figure~\ref{fig:overview}(c) connects this source review to executable verification.

\paragraph{Constructing assessable observations.}
For each data requirement $r$, the evaluation specification identifies the permitted fields $F_r$, the observation to compute, a comparison operator $\omega_r$, and a target value $\gamma_r$. Source review checks whether these fields provide sufficient evidence and whether the observation and comparison preserve the source condition's meaning and strength. Evidence limitations are recorded when a condition cannot be assessed from the retained fields.

Let $R_{\mathrm c}$ and $R_{\mathrm s}$ denote the data requirements assigned to computational and semantic evaluation, respectively. Their observed values are
\begin{equation}
z_r(D)=
\begin{cases}
g_r(D[F_r]), & r\in R_{\mathrm c},\\
\operatorname{Agg}_r\bigl((J_r(e))_{e\in U_r(D[F_r])}\bigr),
    & r\in R_{\mathrm s}.
\end{cases}
\label{eq:observed_value}
\end{equation}
Here $D[F_r]$ restricts the data to the permitted fields, and $g_r$ is a generated computational evaluator. For semantic checks, $U_r$ constructs evidence units $e$ using fixed selection and preparation rules, $J_r$ supplies an LLM judgment for each unit, and $\operatorname{Agg}_r$ combines these judgments using a specified aggregation rule. Empty inputs and unknown judgments follow the evaluator's declared policies.

\paragraph{Producing verdicts and review evidence.}
Within a fixed assessment version $\nu$, whose superscript is omitted here, the runtime computes the verdict for a valid observation as
\begin{equation}
    V_r(D)=\mathbf{1}[z_r(D)\;\omega_r\;\gamma_r].
    \label{eq:runtime_verdict}
\end{equation}
Here $\mathbf{1}[\cdot]$ equals 1 when the comparison holds and 0 otherwise. Generated evaluators supply observations; the runtime applies the reviewed comparison. Program validation checks permitted operations, field references, and output structure. Execution errors and invalid outputs are recorded for review rather than treated as successful checks. Observations, verdicts, errors, and source-review findings are linked to the relevant requirement and evaluator versions, supplying evidence for the correction and acceptance decisions in Section~\ref{sec:adaptive_control}.

\subsection{Diagnostic sample synthesis}
\label{sec:sample_synthesis}

Synthesis agents execute the controller's construction actions across data, visual, and question stages. Each stage returns candidate artifacts and review evidence to the controller, which determines whether to adopt them, continue correction, or revisit a permitted stage.

\paragraph{Data construction.}
Agents generate data under the reviewed field, structure, and difficulty requirements, specifying the entities, attributes, numerical patterns, and relationships needed for the diagnostic target. Deterministic checks examine data validity before semantic evaluation. Failed checks return to the data-repair loop in Section~\ref{sec:adaptive_control}. Once all applicable checks pass, the admitted data $D^{\star}$ become the fixed factual reference for visual and question construction.

\paragraph{Visual realization.}
Agents translate $D^{\star}$ and the visual requirements into a brief and assessable drawing goals. As outlined in Figure~\ref{fig:overview}(a), image generation produces the main visual content, while code provides readable text, exact annotations, and simple marks. Each update is reviewed against its goals using the full image and relevant regions. Review examines factual fidelity, goal fulfillment, and preservation of previously accepted content. Useful partial results may support another iteration without marking their goals complete; damage introduced by a later edit reopens the affected goals and checks.

\begin{table}[!t]
\caption{Semantic deviations during natural construction of 290 samples. Counts and percentages refer to samples with a recorded semantic deviation at each construction stage. A sample can exhibit deviations at several stages. Formatting-only and explanation-only issues are excluded.}
\label{tab:semantic-control}
\centering
\small
\setlength{\tabcolsep}{2pt}
\renewcommand{\arraystretch}{1.08}
\begin{tabular*}{\columnwidth}{@{\extracolsep{\fill}}l>{\raggedright\arraybackslash}p{0.49\columnwidth}r@{}}
\toprule
Stage & Recorded deviation & $n$ (\%) \\
\midrule
Requirements & Source conditions weakened, strengthened, or replaced. & 74 (25.5) \\
\addlinespace[3pt]
Data & Required contrasts, relations, or category meanings altered. & 19 (6.6) \\
\addlinespace[3pt]
Image & Encoding changed, evidence lost, or shortcuts introduced. & 126 (43.4) \\
\addlinespace[3pt]
Anchors & Unsupported evidence or operations that miss the target. & 44 (15.2) \\
\addlinespace[3pt]
QA & Image, target judgment, or required difficulty bypassed. & 139 (47.9) \\
\bottomrule
\end{tabular*}

\end{table}

\paragraph{Diagnostic question construction.}
Mechanism anchors $M$ connect the target error to image regions, supporting facts, and the visual operations a question should require. Once the anchors pass review for factual support, target alignment, and difficulty, agents generate the question $q$, options $O$, and reference answer $a$.

Reviews examine factual correctness, whether the wording alone reveals the answer, whether the intended visual judgment is needed, and difficulty under the simplest supported solution and option-elimination routes. Final review checks the reference answer against the source data and assesses whether the model-visible inputs $(I,q,O)$ support its selection. Findings identify the affected image or question components for the controller's bounded correction and backtracking.

\section{Experiments}
\label{sec:experiments}

\subsection{Experimental settings}

DRIC-Agent uses GPT-6.1-Terra for generation, review, and repair, and GPT Image 2.6 Flare for image synthesis. We evaluate 290 diagnostic samples covering 59 entries in the error taxonomy, organized into four top-level groups: visual perception and encoding, structure and layout, data logic and reasoning, and semantic understanding. The final configurations span 53 scenes and 65 visual types after compatibility checking.

For each sample, all 15 evaluated MLLMs receive the same $(I,q,O)$ and return an option letter with a brief explanation. Accuracy measures agreement with the reference option; explanations are not scored. We report 10 closed-source and 5 open-source models separately. Overall accuracy pools all samples, while accuracy for each top-level group uses its corresponding subset. Means within each model group weight models equally.

Table~\ref{tab:model-accuracy} shows overall accuracies ranging from 55.5\% to 91.0\%. Mean accuracy is 88.5\% for closed-source models and 71.2\% for open-source models. Within-model differences across the four top-level groups range from 5.1 to 18.0 percentage points, revealing capability differences obscured by overall scores. GPT-6.1-Sol leads overall at 91.0\%, whereas Gemini 3.8 Flash and Claude Opus 5.5 share the highest accuracy in structure and layout at 91.7\%. No model leads all four groups, so a single overall ranking does not capture every capability profile. We next examine semantic fidelity during construction and in final samples.

\subsection{Semantic control during construction}
\label{sec:semantic-control}

We inspect intermediate candidates and review records for all 290 samples. A semantic deviation changes target conditions, data relations, visual evidence, anchored operations, or the necessity or difficulty of the intended judgment. Execution, formatting, and explanation-only defects are excluded.

Table~\ref{tab:semantic-control} documents natural deviations without injected faults or additional generation. The records identify intermediate requirements or checks that deviate from source conditions in 25.5\% of samples, and deviations in visual evidence or encoding in 43.4\%. Faulty requirements can change what is checked, image changes can alter the available evidence, and QA shortcuts can bypass the intended judgment. These distinct failure modes motivate reviewing both assessment conditions and generated content. The counts describe intermediate deviations; they do not estimate the causal reduction achieved by correction.

\subsection{Ablation study}
\label{sec:ablation-study}

We examine final diagnostic fidelity, complement model-based judgments with human review, and then test the utility of targeted follow-up construction.

\setcounter{table}{3}
\begin{table}[!t]
\caption{Diagnostic judgments (\%) on 19 paired contexts. Complete fidelity requires answer validity, target conditions, and judgment necessity jointly within each reviewer, then a two-of-three majority. $\Delta$: original target fidelity minus complete fidelity (percentage points), calculated before rounding.}
\label{tab:visual-qa-fidelity}
\centering
\small
\setlength{\tabcolsep}{1.5pt}
\renewcommand{\arraystretch}{1.08}
\begin{tabular*}{\columnwidth}{@{\extracolsep{\fill}}lccccc@{}}
\toprule
& \shortstack{Original\\audit} & \multicolumn{3}{c}{Solution-aware audit} & \\
\cmidrule(lr){2-2}\cmidrule(lr){3-5}
Method & \shortstack{Target\\fidelity} & \shortstack{Answer\\validity} & \shortstack{Judgment\\necessity} & \shortstack{Complete\\target\\fidelity} & $\Delta$ \\
\midrule
Direct & 84.2 & 78.9 & 63.2 & 47.4 & 36.8 \\
DRIC-Agent & 94.7 & 100.0 & 78.9 & 73.7 & 21.1 \\
\bottomrule
\end{tabular*}

\end{table}

\paragraph{Diagnostic fidelity under complementary audits.}

For each paired context, Direct receives the same source data, error-taxonomy specification, scene, visual type, and difficulty requirements as DRIC-Agent. It uses the same construction models to generate one image and one QA item without revision; DRIC-Agent contributes its existing final sample.

The original audit assesses target conditions and question alignment separately from judgment necessity. In the solution-aware audit, GPT-6.1-Sol, Gemini 4 Carbon, and Claude Opus 5.5 first solve anonymized items without diagnostic target specifications or answer keys. They then assess answer validity, target conditions, and judgment necessity using their solutions, the targets, and source facts. Author explanations and construction reviews are withheld.

In Table~\ref{tab:visual-qa-fidelity}, original fidelity exceeds complete fidelity by 36.8 percentage points for Direct and 21.1 for DRIC-Agent. Visible target conditions can coexist with an invalid answer or a shortcut around the intended judgment. Both the inputs and acceptance criteria change between audits, so the gap cannot be attributed solely to including solution evidence. Complete fidelity is 73.7\% for DRIC-Agent versus 47.4\% for Direct. This workflow comparison complements Table~\ref{tab:semantic-control}.

\setcounter{table}{4}
\begin{table}[t]
\caption{Human judgments on 59 final samples (\% positive). Four reviewers assess the same samples, one per taxonomy entry from the 290-item set. Mean pools all four reviewers.}
\label{tab:human-audit}
\centering
\small
\setlength{\tabcolsep}{2pt}
\renewcommand{\arraystretch}{1.08}
\begin{tabular*}{\columnwidth}{@{\extracolsep{\fill}}lccccc@{}}
\toprule
Criterion & R1 & R2 & R3 & R4 & Mean \\
\midrule
Answer validity & 89.8 & 89.8 & 94.9 & 69.5 & 86.0 \\
Target conditions & 81.4 & 91.5 & 84.7 & 76.3 & 83.5 \\
Judgment necessity & 100.0 & 98.3 & 100.0 & 100.0 & 99.6 \\
\bottomrule
\end{tabular*}

\end{table}

\paragraph{Human evaluation of diagnostic fidelity.}
\label{sec:human-evaluation}

An MLLM's solution may not capture all valid human solution routes. Four human reviewers therefore assess 59 samples from the 290-item set, one per taxonomy entry, selected without consulting model answers or human ratings. Reviewers inspect images, questions, options, reference answers and explanations, and error-taxonomy specifications, with Chinese translations. The reference answers themselves are also subject to review.

Across 236 assessments, answer validity, target-condition preservation, and judgment necessity receive 86.0\%, 83.5\%, and 99.6\% positive judgments, respectively (Table~\ref{tab:human-audit}). Mean pairwise agreement is 82.8\%, 76.6\%, and 99.2\%, respectively. The criteria assess distinct properties; a positive rating on one does not establish the others.

\setcounter{table}{5}
\begin{table}[t]
\caption{Error-prioritized follow-up subsets (8 samples per taxonomy entry). Samples are selected after evaluation, prioritizing errors by both models, then one, then neither. Target: subset accuracy (\%); L1: the same model's accuracy for the corresponding top-level group in Table~\ref{tab:model-accuracy}; $\Delta=\mathrm{L1}-\mathrm{Target}$ (percentage points).}
\label{tab:targeted-generation}
\centering
\small
\setlength{\tabcolsep}{2pt}
\renewcommand{\arraystretch}{1.05}
\begin{tabular*}{\columnwidth}{@{\extracolsep{\fill}}lrrr@{}}
\toprule
Model & L1 & Target & $\Delta$ \\
\midrule
\multicolumn{4}{@{}l}{\textit{Color gradient}} \\
Gemini 4 Carbon & 90.0 & 62.5 & 27.5 \\
GLM-5.2 & 88.8 & 62.5 & 26.3 \\
\addlinespace[3pt]
\multicolumn{4}{@{}l}{\textit{Stacked elements}} \\
DeepSeek V4.1 Flash & 53.8 & 25.0 & 28.8 \\
GPT-5.6-Sol & 82.5 & 50.0 & 32.5 \\
\addlinespace[3pt]
\multicolumn{4}{@{}l}{\textit{Dual Y axes}} \\
GPT-6.1-Sol & 88.3 & 62.5 & 25.8 \\
Gemini 3.8 Flash & 91.7 & 75.0 & 16.7 \\
\bottomrule
\end{tabular*}

\end{table}

\paragraph{Error-guided follow-up construction.}
\label{sec:targeted-generation}

After evaluating the 290 samples, we generated additional samples for error-taxonomy entries on which specific models had shown weaknesses. Table~\ref{tab:targeted-generation} presents this targeted follow-up for three entries, using the same eight samples for each model pair.

\section{Limitations}
\label{sec:limitations}
Human reviewers and MLLMs may differ in their review practices and visual preferences. Automated aesthetic judgments may not match human preferences, and generated questions can retain verbose phrasing that makes human interpretation harder. We have not isolated the effects of visual type on model responses or systematically analyzed the mechanisms underlying individual models' errors. QA-level drift checks also depend on the solution routes considered during review. Future work will isolate visual-type effects, investigate model-specific errors, and test for valid solution routes overlooked by MLLM critics.

\section{Conclusion}
\label{sec:conclusion}
We presented DRIC-Agent, a framework for controlling diagnostic drift in visual benchmark construction from sparse error-taxonomy specifications. Source review directs corrections to content, evaluators, or requirements, while acceptance rules distinguish data repair under fixed criteria from changes to those criteria. Separating source fidelity from agreement with current checks addresses drift that can arise when an inferred requirement is itself flawed. Re-evaluating the current data after an authorized assessment change establishes an explicit baseline for subsequent repair decisions. Evaluation reveals differences in model performance and documents semantic deviations during construction. Paired audits report higher complete target fidelity than direct construction, while human review provides complementary judgments about final samples. Targeted follow-up cases illustrate how observed weaknesses can guide further testing. Retaining the source target and evidence behind each revision makes construction decisions inspectable throughout the workflow. This links diagnostic targets to the requirements, visual evidence, and judgments.

\section*{LLM Usage Statement}
OpenAI Codex assisted with manuscript drafting and revision, literature organization, analysis and visualization code, and the human-audit interface; GPT Image 2 assisted with schematic icons in Figure~\ref{fig:overview}. The authors retain responsibility for the study design, factual and bibliographic accuracy, analyses, and final manuscript.
{\small
\bibliographystyle{ieeenat_fullname}
\bibliography{references}

@article{gebru2021datasheets,
  title = {{Datasheets for Datasets}},
  author = {Gebru, Timnit and Morgenstern, Jamie and Vecchione, Briana and Vaughan, Jennifer Wortman and Wallach, Hanna and Daum{\'e}, III, Hal and Crawford, Kate},
  journal = {Communications of the ACM},
  volume = {64},
  number = {12},
  pages = {86--92},
  year = {2021},
  url = {https://doi.org/10.1145/3458723},
  doi = {10.1145/3458723},
}

@inproceedings{bean2025constructvalidity,
  title = {{Measuring what Matters: Construct Validity in Large Language Model Benchmarks}},
  author = {Bean, Andrew M. and Kearns, Ryan Othniel and Romanou, Angelika and Hafner, Franziska Sofia and Mayne, Harry and Batzner, Jan and Foroutan, Negar and Schmitz, Chris and Korgul, Karolina and Batra, Hunar and Deb, Oishi and Beharry, Emma and Emde, Cornelius and Foster, Thomas and Gausen, Anna and Grandury, Mar{\'i}a and Han, Simeng and Hofmann, Valentin and Ibrahim, Lujain and Kim, Hazel and Kirk, Hannah Rose and Lin, Fangru and Liu, Gabrielle Kaili-May and Luettgau, Lennart and Magomere, Jabez and Rystr{\o}m, Jonathan and Sotnikova, Anna and Yang, Yushi and Zhao, Yilun and Bibi, Adel and Bosselut, Antoine and Clark, Ronald and Cohan, Arman and Foerster, Jakob and Gal, Yarin and Hale, Scott A. and Raji, Inioluwa Deborah and Summerfield, Christopher and Torr, Philip H. S. and Ududec, Cozmin and Rocher, Luc and Mahdi, Adam},
  year = {2025},
  booktitle = {Advances in Neural Information Processing Systems},
  volume = {38},
  doi = {10.52202/085713-0590},
  url = {https://proceedings.neurips.cc/paper_files/paper/2025/hash/1967e0fc3aa6cbbace562f5cb8e3954e-Abstract-Datasets_and_Benchmarks_Track.html},
}

@inproceedings{li2025autobencher,
  title = {{AutoBencher: Towards Declarative Benchmark Construction}},
  author = {LI, XIANG and Kaiyom, Farzaan and Liu, Evan and Mai, Yifan and Liang, Percy and Hashimoto, Tatsunori},
  year = {2025},
  booktitle = {International Conference on Learning Representations},
  volume = {2025},
  pages = {94357--94386},
  url = {https://proceedings.iclr.cc/paper_files/paper/2025/hash/eb216114f3eaad22506fd1bc7bbff0ca-Abstract-Conference.html},
}

@article{bao2024autobenchv,
  title = {{AutoBench-V: Can Large Vision-Language Models Benchmark Themselves?}},
  author = {Bao, Han and Huang, Yue and Wang, Yanbo and Ye, Jiayi and Wang, Xiangqi and Chen, Xiuying and Zhao, Yue and Zhou, Tianyi and Elhoseiny, Mohamed and Zhang, Xiangliang},
  journal = {arXiv preprint arXiv:2410.21259},
  year = {2024},
  url = {https://arxiv.org/abs/2410.21259v4},
}

@article{kondic2025chartgen,
  title = {{ChartGen: Scaling Chart Understanding via Code-Guided Synthetic Chart Generation}},
  author = {Kondic, Jovana and Li, Pengyuan and Joshi, Dhiraj and He, Zexue and Abedin, Shafiq and Sun, Jennifer and Wiesel, Ben and Schwartz, Eli and Nassar, Ahmed and Wu, Bo and Arbelle, Assaf and Oliva, Aude and Gutfreund, Dan and Karlinsky, Leonid and Feris, Rogerio},
  journal = {arXiv preprint arXiv:2507.19492},
  year = {2025},
  url = {https://arxiv.org/abs/2507.19492v1},
}

@inproceedings{johnson2017clevr,
  title = {{CLEVR}: A Diagnostic Dataset for Compositional Language and Elementary Visual Reasoning},
  author = {Johnson, Justin and Hariharan, Bharath and van der Maaten, Laurens and Fei-Fei, Li and Zitnick, C. Lawrence and Girshick, Ross},
  booktitle = {Proceedings of the IEEE Conference on Computer Vision and Pattern Recognition},
  pages = {2901--2910},
  year = {2017},
  url = {https://openaccess.thecvf.com/content_cvpr_2017/html/Johnson_CLEVR_A_Diagnostic_CVPR_2017_paper.html},
}

@inproceedings{zhang2024taskmeanything,
  title = {{Task Me Anything}},
  author = {Zhang, Jieyu and Huang, Weikai and Ma, Zixian and Michel, Oscar and He, Dong and Gupta, Tanmay and Ma, Wei-Chiu and Farhadi, Ali and Kembhavi, Aniruddha and Krishna, Ranjay},
  year = {2024},
  booktitle = {Advances in Neural Information Processing Systems},
  volume = {37},
  pages = {19965--19974},
  doi = {10.52202/079017-0629},
  url = {https://proceedings.neurips.cc/paper_files/paper/2024/hash/237ffa9a473eff1c66d085dba7f813ba-Abstract-Datasets_and_Benchmarks_Track.html},
}

@inproceedings{he2025reachqa,
  title = {{Distill Visual Chart Reasoning Ability from LLMs to MLLMs}},
  author = {He, Wei  and
      Xi, Zhiheng  and
      Zhao, Wanxu  and
      Fan, Xiaoran  and
      Ding, Yiwen  and
      Shan, Zifei  and
      Gui, Tao  and
      Zhang, Qi  and
      Huang, Xuanjing},
  booktitle = {Findings of the {A}ssociation for {C}omputational {L}inguistics: {EMNLP} 2025},
  year = {2025},
  publisher = {Association for Computational Linguistics},
  url = {https://aclanthology.org/2025.findings-emnlp.172/},
  doi = {10.18653/v1/2025.findings-emnlp.172},
  pages = {3224--3250},
}

@inproceedings{liu2026chartverse,
  title = {{ChartVerse: Scaling Chart Reasoning via Reliable Programmatic Synthesis from Scratch}},
  author = {Liu, Zheng  and
      Lin, Honglin  and
      Wang, Xiaoyang  and
      Gao, Xin  and
      Li, Yu  and
      Cai, Mengzhang  and
      Zhu, Yun  and
      Zhong, Zhanping  and
      Pei, Qizhi  and
      Pan, Zhuoshi  and
      Shang, Xiaoran  and
      He, Conghui  and
      Cui, Bin  and
      Zhang, Wentao  and
      Wu, Lijun},
  booktitle = {Proceedings of the 64th Annual Meeting of the {A}ssociation for {C}omputational {L}inguistics (Volume 1: Long Papers)},
  year = {2026},
  publisher = {Association for Computational Linguistics},
  url = {https://aclanthology.org/2026.acl-long.344/},
  doi = {10.18653/v1/2026.acl-long.344},
  pages = {7551--7577},
}

@article{chen2026bloomqa,
  title = {{Automated Benchmark Generation from Domain Guidelines Informed by Bloom's Taxonomy}},
  author = {Chen, Si and Khiem, Le Huy and Szymanski, Annalisa and Metoyer, Ronald and Hua, Ting and Chawla, Nitesh V.},
  journal = {arXiv preprint arXiv:2601.20253},
  year = {2026},
  url = {https://arxiv.org/abs/2601.20253v1},
}

@inproceedings{sokol2025benchmarkcards,
  title = {{BenchmarkCards: Standardized Documentation for Large Language Model Benchmarks}},
  author = {Sokol, Anna and Daly, Elizabeth and Hind, Michael and Piorkowski, David and Zhang, Xiangliang and Moniz, Nuno and Chawla, Nitesh},
  year = {2025},
  booktitle = {Advances in Neural Information Processing Systems},
  volume = {38},
  doi = {10.52202/085713-2748},
  url = {https://proceedings.neurips.cc/paper_files/paper/2025/hash/76175f4355e2f67cf91be468c8860070-Abstract-Datasets_and_Benchmarks_Track.html},
}

@inproceedings{butt2024automatedbenchmark,
  author = {Butt, Natasha and Chandrasekaran, Varun and Joshi, Neel and Nushi, Besmira and Balachandran, Vidhisha},
  title = {{BenchAgents: Automated Benchmark Creation with Agent Interaction}},
  booktitle = {2nd Workshop on Navigating and Addressing Data Problems for Foundation Models at ICLR},
  year = {2025},
  url = {https://openreview.net/forum?id=Xh6S3X3enu},
}

@article{li2026benchmarkagent,
  title = {{Benchmark Everything Everywhere All at Once}},
  author = {Xiong, Shiyun and Wu, Dongming and Sun, Peiwen and Ai, Yuang and Yang, Bokang and Han, Wencheng and Li, Xiao-Hui and Yue, Xiangyu},
  journal = {arXiv preprint arXiv:2606.06462},
  year = {2026},
  url = {https://arxiv.org/abs/2606.06462v1},
}

@article{tang2026rankjudge,
  title = {{RankJudge: A Multi-Turn LLM-as-a-Judge Synthetic Benchmark Generator}},
  author = {Tang, Zhenwei and Liu, Zhaoyan and Hosseinzadeh, Rasa and Wu, Tongzi and Golestan, Keyvan and Cresswell, Jesse C.},
  journal = {arXiv preprint arXiv:2605.21748},
  year = {2026},
  url = {https://arxiv.org/abs/2605.21748v1},
}

@article{jiang2026embodiedbenchclaw,
  title = {{Embodied-BenchClaw: An Autonomous Multi-Agent System for Embodied Spatial Intelligence Benchmark Construction}},
  author = {Jiang, Baoyang and Zhang, Fengchun and Wang, Leyuan and Li, Haotian and Wang, Yida and Ji, Zhe and Lai, Jinshan and Ren, Xi and Hu, Jianwei and Ma, Qiang},
  journal = {arXiv preprint arXiv:2606.11909},
  year = {2026},
  url = {https://arxiv.org/abs/2606.11909v1},
}

@inproceedings{huang2026probellm,
  title = {{ProbeLLM: Automating Principled Diagnosis of LLM Failures}},
  author = {Yue Huang and Zhengzhe Jiang and Yuchen Ma and Yu Jiang and Xiangqi Wang and Yujun Zhou and Yuexing Hao and Kehan Guo and Pin-Yu Chen and Marzyeh Ghassemi and Stefan Feuerriegel and Xiangliang Zhang},
  year = {2026},
  booktitle = {Proceedings of the 43rd International Conference on Machine Learning},
  pages = {46721--46751},
  url = {https://proceedings.mlr.press/v306/huang26r.html},
  series = {Proceedings of Machine Learning Research},
  volume = {306},
  publisher = {PMLR},
}

@article{ivanov2026anchor,
  title = {{Anchor: Preventing Artifact Drift for Reliable Agent Verification}},
  author = {Rana, Abhijay and Ivanov, Maksim},
  journal = {arXiv preprint arXiv:2605.26321},
  year = {2026},
  url = {https://arxiv.org/abs/2605.26321v2},
}

@article{ranganath2026pipecypher,
  title = {{PIPE-Cypher: Automatic Enterprise Benchmark Generation for Text-to-Cypher Systems}},
  author = {Ranganath, Suraj and Raghavendra, Anish},
  journal = {arXiv preprint arXiv:2606.08481},
  year = {2026},
  url = {https://arxiv.org/abs/2606.08481v1},
}

@article{li2026targetedtests,
  title = {{Targeted Tests for LLM Reasoning: An Audit-Constrained Protocol}},
  author = {Li, Hongmin},
  journal = {arXiv preprint arXiv:2605.11599},
  year = {2026},
  url = {https://arxiv.org/abs/2605.11599v3},
}

@article{chen2025arenabencher,
  title = {{ArenaBencher: Automatic Benchmark Evolution via Multi-Model Competitive Evaluation}},
  author = {Liu, Qin and Dineen, Jacob and Huang, Yuxi and Zhang, Sheng and Poon, Hoifung and Zhou, Ben and Chen, Muhao},
  journal = {arXiv preprint arXiv:2510.08569},
  year = {2025},
  url = {https://arxiv.org/abs/2510.08569v1},
}

@inproceedings{sun2024autohallusion,
  title = {{AutoHallusion: Automatic Generation of Hallucination Benchmarks for Vision-Language Models}},
  author = {Wu, Xiyang  and
      Guan, Tianrui  and
      Li, Dianqi  and
      Huang, Shuaiyi  and
      Liu, Xiaoyu  and
      Wang, Xijun  and
      Xian, Ruiqi  and
      Shrivastava, Abhinav  and
      Huang, Furong  and
      Boyd-Graber, Jordan  and
      Zhou, Tianyi  and
      Manocha, Dinesh},
  booktitle = {Findings of the {A}ssociation for {C}omputational {L}inguistics: {EMNLP} 2024},
  year = {2024},
  publisher = {Association for Computational Linguistics},
  url = {https://aclanthology.org/2024.findings-emnlp.493/},
  doi = {10.18653/v1/2024.findings-emnlp.493},
  pages = {8395--8419},
}

@inproceedings{liu2026chartnet,
  title = {{ChartNet: A Million-Scale, High-Quality Multimodal Dataset for Robust Chart Understanding}},
  author = {Kondic, Jovana and Li, Pengyuan and Joshi, Dhiraj and Sanchez, Isaac and Wiesel, Ben and Abedin, Shafiq and Alfassy, Amit and Schwartz, Eli and Caraballo, Daniel and Cinar, Yagmur Gizem and Scheidegger, Florian and Ross, Steven I. and Weidele, Daniel Karl I. and Hua, Hang and Arutyunova, Ekaterina and Herzig, Roei and Wang, Zihan and Yu, Xinyue and Zhao, Yunfei and Jiang, Sicong and Liu, Minghao and Lin, Qunshu and Oliva, Aude and Feris, Rogerio},
  year = {2026},
  booktitle = {Proceedings of the IEEE/CVF Conference on Computer Vision and Pattern Recognition},
  pages = {15922--15932},
  url = {https://openaccess.thecvf.com/content/CVPR2026/html/Kondic_ChartNet_A_Million-Scale_High-Quality_Multimodal_Dataset_for_Robust_Chart_Understanding_CVPR_2026_paper.html},
}

@inproceedings{wang2025chartm3,
  title = {{ChartM$^3$: A Multi-Stage Code-Driven Pipeline for Constructing Multi-Dimensional and Multi-Step Visual Reasoning Data in Chart Comprehension}},
  author = {Xu, Duo  and
      Cheng, Hao  and
      Lin, Xin  and
      Xie, Zhen  and
      Wang, Hao Henry},
  booktitle = {Findings of the {A}ssociation for {C}omputational {L}inguistics: {EMNLP} 2025},
  year = {2025},
  publisher = {Association for Computational Linguistics},
  url = {https://aclanthology.org/2025.findings-emnlp.701/},
  doi = {10.18653/v1/2025.findings-emnlp.701},
  pages = {13046--13068},
}

@inproceedings{liu2024ecbd,
  title = {{ECBD: Evidence-Centered Benchmark Design for NLP}},
  author = {Liu, Yu Lu  and
      Blodgett, Su Lin  and
      Cheung, Jackie  and
      Liao, Q. Vera  and
      Olteanu, Alexandra  and
      Xiao, Ziang},
  booktitle = {Proceedings of the 62nd Annual Meeting of the {A}ssociation for {C}omputational {L}inguistics (Volume 1: Long Papers)},
  year = {2024},
  publisher = {Association for Computational Linguistics},
  url = {https://aclanthology.org/2024.acl-long.861/},
  doi = {10.18653/v1/2024.acl-long.861},
  pages = {16349--16365},
}

@inproceedings{diddee2026benchbrowser,
  title = {{BenchBrowser: Retrieving Evidence for Evaluating Benchmark Validity}},
  author = {Diddee, Harshita  and
      Yauney, Gregory  and
      Swayamdipta, Swabha  and
      Ippolito, Daphne},
  booktitle = {Proceedings of the 2026 Conference on Empirical Methods in Natural Language Processing},
  year = {2026},
  publisher = {Association for Computational Linguistics},
  url = {https://aclanthology.org/2026.emnlp-main.1755/},
  doi = {10.18653/2026.emnlp-main.1755},
  pages = {40024--40053},
}

@inproceedings{jiang2026itemlevel,
  author = {Jiang, Han and Zhang, Susu and Zhu, Dongyao and Bai, Yuzhuo and Truong, Sang T. and Yi, Xiaoyuan and Koyejo, Sanmi and Xie, Xing and Xiao, Ziang},
  title = {{AI Evaluation Should Require Standardized Item-Level Data Releases}},
  booktitle = {Advances in Neural Information Processing Systems},
  year = {2026},
  url = {https://neurips.cc/virtual/2026/poster/140079},
  note = {Accepted, Position Paper; final proceedings metadata pending},
}

@inproceedings{madaan2023selfrefine,
  title = {{Self-Refine: Iterative Refinement with Self-Feedback}},
  author = {Madaan, Aman and Tandon, Niket and Gupta, Prakhar and Hallinan, Skyler and Gao, Luyu and Wiegreffe, Sarah and Alon, Uri and Dziri, Nouha and Prabhumoye, Shrimai and Yang, Yiming and Gupta, Shashank and Majumder, Bodhisattwa Prasad and Hermann, Katherine and Welleck, Sean and Yazdanbakhsh, Amir and Clark, Peter},
  year = {2023},
  booktitle = {Advances in Neural Information Processing Systems},
  volume = {36},
  pages = {46534--46594},
  doi = {10.52202/075280-2019},
  url = {https://proceedings.neurips.cc/paper_files/paper/2023/hash/91edff07232fb1b55a505a9e9f6c0ff3-Abstract-Conference.html},
}

@inproceedings{shinn2023reflexion,
  title = {{Reflexion: language agents with verbal reinforcement learning}},
  author = {Shinn, Noah and Cassano, Federico and Gopinath, Ashwin and Narasimhan, Karthik and Yao, Shunyu},
  year = {2023},
  booktitle = {Advances in Neural Information Processing Systems},
  volume = {36},
  pages = {8634--8652},
  doi = {10.52202/075280-0377},
  url = {https://proceedings.neurips.cc/paper_files/paper/2023/hash/1b44b878bb782e6954cd888628510e90-Abstract-Conference.html},
}

@inproceedings{dhuliawala2024chainverification,
  title = {{Chain-of-Verification Reduces Hallucination in Large Language Models}},
  author = {Dhuliawala, Shehzaad  and
      Komeili, Mojtaba  and
      Xu, Jing  and
      Raileanu, Roberta  and
      Li, Xian  and
      Celikyilmaz, Asli  and
      Weston, Jason},
  booktitle = {Findings of the {A}ssociation for {C}omputational {L}inguistics: {ACL} 2024},
  year = {2024},
  publisher = {Association for Computational Linguistics},
  url = {https://aclanthology.org/2024.findings-acl.212/},
  doi = {10.18653/v1/2024.findings-acl.212},
  pages = {3563--3578},
}

@inproceedings{geng2023grammar,
  title = {{Grammar-Constrained Decoding for Structured NLP Tasks without Finetuning}},
  author = {Geng, Saibo  and
      Josifoski, Martin  and
      Peyrard, Maxime  and
      West, Robert},
  booktitle = {Proceedings of the 2023 Conference on Empirical Methods in Natural Language Processing},
  year = {2023},
  publisher = {Association for Computational Linguistics},
  url = {https://aclanthology.org/2023.emnlp-main.674/},
  doi = {10.18653/v1/2023.emnlp-main.674},
  pages = {10932--10952},
}

@misc{geng2025jsonschema,
  author = {Geng, Saibo and Cooper, Hudson and Moskal, Michal and Jenkins, Samuel and Berman, Julian and Ranchin, Nathan and West, Robert and Horvitz, Eric and Nori, Harsha},
  title = {{JSONSchemaBench: Evaluating Constrained Decoding with LLMs on Efficiency, Coverage and Quality}},
  howpublished = {ES-FoMo III: Efficient Systems for Foundation Models Workshop at ICML},
  year = {2025},
  note = {Non-archival presentation},
  url = {https://openreview.net/forum?id=FKOaJqKoio},
}

@article{simbola2026compliance,
  title = {{LLM-Driven Compliance Checking for Natural-Language Policies over Data Product Descriptors}},
  author = {Simbola, Francesco and Reforgiato Recupero, Diego and Riboni, Daniele and Salis, Martina},
  year = {2026},
  journal = {Machine Learning},
  volume = {115},
  number = {5},
  eid = {116},
  doi = {10.1007/s10994-026-07038-6},
  url = {https://link.springer.com/article/10.1007/s10994-026-07038-6},
}

@inproceedings{yang2025cosyn,
  title = {{Scaling Text-Rich Image Understanding via Code-Guided Synthetic Multimodal Data Generation}},
  author = {Yang, Yue  and
      Patel, Ajay  and
      Deitke, Matt  and
      Gupta, Tanmay  and
      Weihs, Luca  and
      Head, Andrew  and
      Yatskar, Mark  and
      Callison-Burch, Chris  and
      Krishna, Ranjay  and
      Kembhavi, Aniruddha  and
      Clark, Christopher},
  booktitle = {Proceedings of the 63rd Annual Meeting of the {A}ssociation for {C}omputational {L}inguistics (Volume 1: Long Papers)},
  year = {2025},
  publisher = {Association for Computational Linguistics},
  url = {https://aclanthology.org/2025.acl-long.855/},
  doi = {10.18653/v1/2025.acl-long.855},
  pages = {17486--17505},
}

@inproceedings{hudson2019gqa,
  title = {{GQA}: A New Dataset for Real-World Visual Reasoning and Compositional Question Answering},
  author = {Hudson, Drew A. and Manning, Christopher D.},
  booktitle = {Proceedings of the IEEE/CVF Conference on Computer Vision and Pattern Recognition},
  pages = {6700--6709},
  year = {2019},
  url = {https://openaccess.thecvf.com/content_CVPR_2019/html/Hudson_GQA_A_New_Dataset_for_Real-World_Visual_Reasoning_and_Compositional_CVPR_2019_paper.html},
}

@inproceedings{chen2024mllmjudge,
  title = {{MLLM}-as-a-Judge: Assessing Multimodal {LLM}-as-a-Judge with Vision-Language Benchmark},
  author = {Chen, Dongping and Chen, Ruoxi and Zhang, Shilin and Wang, Yaochen and Liu, Yinuo and Zhou, Huichi and Zhang, Qihui and Wan, Yao and Zhou, Pan and Sun, Lichao},
  booktitle = {Proceedings of the 41st International Conference on Machine Learning},
  series = {Proceedings of Machine Learning Research},
  volume = {235},
  pages = {6562--6595},
  year = {2024},
  publisher = {PMLR},
  url = {https://proceedings.mlr.press/v235/chen24h.html},
}
}
\end{document}